\documentclass[10pt,twocolumn,letterpaper]{article}

\usepackage{iccv}
\usepackage{times}
\usepackage{epsfig}
\usepackage{graphicx}
\usepackage{amsmath}
\usepackage{amssymb}
\usepackage{multirow}
\usepackage{tabularx}

\usepackage[breaklinks=true,bookmarks=false]{hyperref}

\iccvfinalcopy 

\def\iccvPaperID{12318} 

\ificcvfinal\fi

\begin{document}

\title{Efficient neural supersampling on a novel gaming dataset}

\author{Antoine Mercier \hspace{0.2em} Ruan Erasmus \hspace{0.2em}  
 Yashesh Savani \hspace{0.2em} Manik Dhingra \hspace{0.2em} Fatih Porikli \hspace{0.2em} Guillaume Berger \\
Qualcomm AI Research\thanks{Qualcomm AI Research is an initiative of Qualcomm Technologies, Inc. and/or its subsidiaries.}\\
{\tt\small \{amercier, rerasmus, ysavani, manidhin, fporikli, guilberg\}@qti.qualcomm.com}
}

\maketitle

\begin{abstract}

Real-time rendering for video games has become increasingly challenging due to the need for higher resolutions, framerates and photorealism. Supersampling has emerged as an effective solution to address this challenge. Our work introduces a novel neural algorithm for supersampling rendered content that is $4\times$ more efficient than existing methods while maintaining the same level of accuracy. Additionally, we introduce a new dataset
 which provides auxiliary modalities such as motion vectors and depth generated using graphics rendering features like viewport jittering and mipmap biasing at different resolutions. We believe that this dataset fills a gap in the current dataset landscape and can serve as a valuable resource to help measure progress in the field and advance the state-of-the-art in super-resolution techniques for gaming content.


\end{abstract}

\section{Introduction}



Real-time rendering has become increasingly difficult for video games due to the demand for higher resolutions, framerates and photorealism. One solution that has recently emerged to address this challenge consists in rendering at lower resolution and then use an upscaling technique to achieve the desired resolution. However, developing efficient upscaling solutions that balance speed and accuracy remains a challenge. Recently, several commercial solutions have been developed for gaming super-resolution, including those that are based on deep learning (DL) such as Nvidia's DLSS \cite{liu2020dlss} or Intel's XeSS \cite{chowdhury2022intel}, as well as solutions that do not rely on machine learning, such as AMD's FSR \cite{fsr1, fsr2_2022}.  Despite the availability of these commercial solutions, there has been relatively little published research on the application of DL-based super-resolution for gaming. We believe that one of the reasons why DL-based super-resolution for gaming has received little attention compared to super-resolution of natural content is that there is currently no standard, publicly available dataset for developing gaming-specific super-resolution solutions. Researchers and developers who want to study or improve upon existing methods must create their own datasets, which can be a time-consuming and resource-intensive process. 


\begin{figure}[t!]
\begin{center}
   \includegraphics[width=\linewidth]{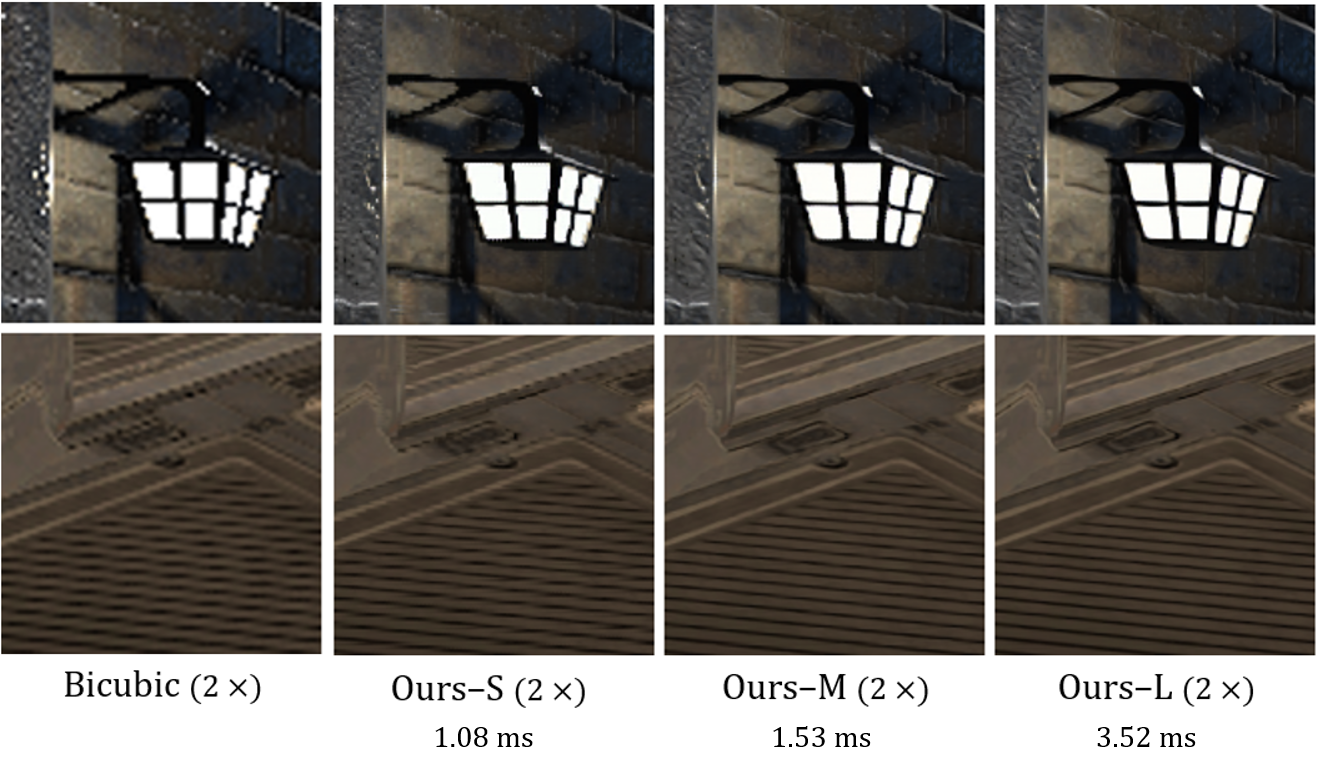}

\caption{Example images produced by our solution using neural networks of different sizes. These models produce 1080p outputs in respectively $1.08$ ms, $1.53$ ms, and $3.52$ ms on an RTX 3090, which is $4\times$ to $12\times$ faster than previous work by Xiao \etal \cite{xiao2020neural}.}
\end{center}
\end{figure}

\begin{figure}[t!]
\begin{center}
   \includegraphics[width=\linewidth]{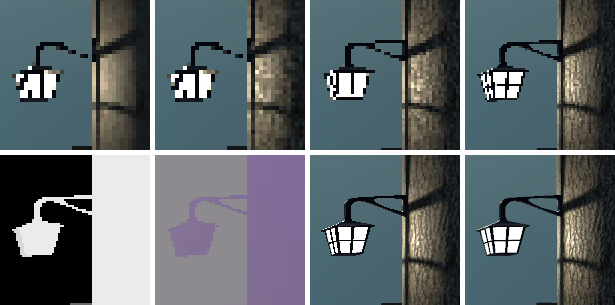}
\caption{Example of data modalities available in the QRISP dataset. \textit{First row, from left to right:} Native 270p, Negative 2 mipmap biased 270p, Negative 1.58 mipmap biased 360p, Negative 1 mipmap biased 540p. \textit{Second row, from left to right:} 540p depth, 540p motion vectors, Native 1080p, Enhanced 1080p}
\label{fig:modalities}
\end{center}
\vspace{-0.5cm}
\end{figure}

\begin{figure*}[!ht]
\begin{center}
   \includegraphics[width=0.95\linewidth]{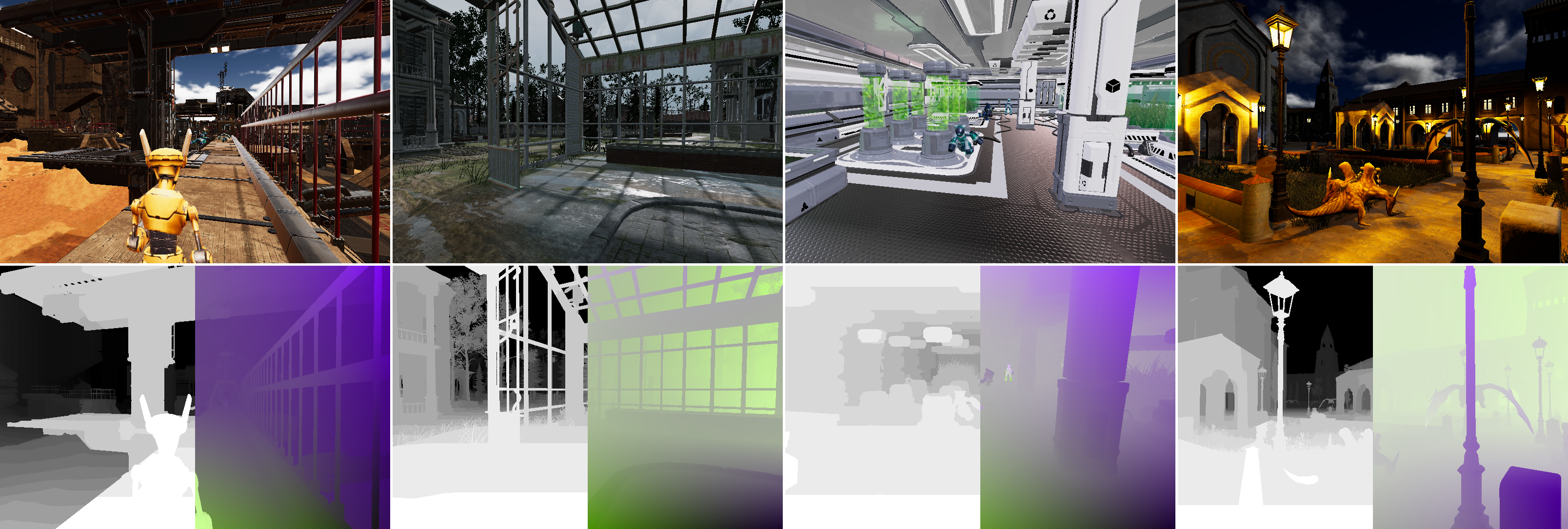}
   \caption{Example of color, depth and motion vector images from the QRISP dataset. In total, our dataset contains 8760 frames (7260 for training, 1500 for testing) from 13 distinct scenes, rendered at different resolutions ranging from 270p to 1080p. More details can be found in the supplementary materials.}
    \label{fig:dataset}
\end{center}
\vspace{-0.5cm}
\end{figure*}

 
Our work makes the following contributions:

\begin{itemize}
    \item we release a dataset specifically designed for the research and development of gaming super-resolution algorithms. We show that models trained on this dataset can compete and outperform the quality levels obtained by commercial solutions such as DLSS \cite{liu2020dlss}. 
    \item we propose an efficient gaming super-resolution architecture which leverages auxiliary modalities (sub-pixel accurate motion vectors, depth) and graphics rendering features (viewport jittering, mipmap biasing) commonly-used for temporal anti-aliasing. Our solution is $4\times$ more efficient than previous published work \cite{xiao2020neural} for the same level of accuracy.
\end{itemize}


Overall, we believe that this work provides a new resource to measure progress in the field and help advance the state-of-the-art in gaming super-resolution.

\section{Related work}

\paragraph{Generic super-resolution.} In recent years,  DL-based approaches for super-resolution of natural content have become increasingly popular \cite{srcnn, fsrcnn, espcn, imdn, hbpn, edsr, rfdn, sr3}, yielding state-of-the-art visual quality compared to interpolation and other algorithmic solutions. In this work, we focus mainly on approaches that exploit information gathered from consecutive frames, as multi-frame super-resolution (also called temporal supersampling in the gaming field) has become the de facto standard for video gaming \cite{liu2020dlss, chowdhury2022intel, fsr2_2022}. Specifically, we consider online super-resolution architectures that can be efficiently stepped forward, as offline video enhancement approaches based on bidirectional mechanisms \cite{huang2015bidirectional, singh2016multi, chan2021basicvsr, liang2022vrt} or sliding windows of input frames \cite{caballero2017real, wang2019edvr, tian2020tdan, li2020mucan} are not suitable for gaming applications. Efficient online multi-frame super-resolution is often based on recurrent convolutional architectures, either with explicit motion compensation \cite{sajjadi2018frame} or without \cite{fuoli2019efficient, isobe2020revisiting, isobe2020video}. Alternatives to explicit motion compensation include architectures based on deformable convolutions \cite{wang2019edvr}, transformers \cite{shi2022rethinking, abati2021efficient, liang2022recurrent} or dynamic upsampling filters \cite{jo2018deep}. In gaming, however, explicit motion compensation is usually preferred, as the game engine can provide precise motion vectors and the neural network can be made much smaller if it doesn't have to learn how to copy past information over long distances.



\begin{figure*}[!ht]
\begin{center}
   \includegraphics[width=\linewidth]{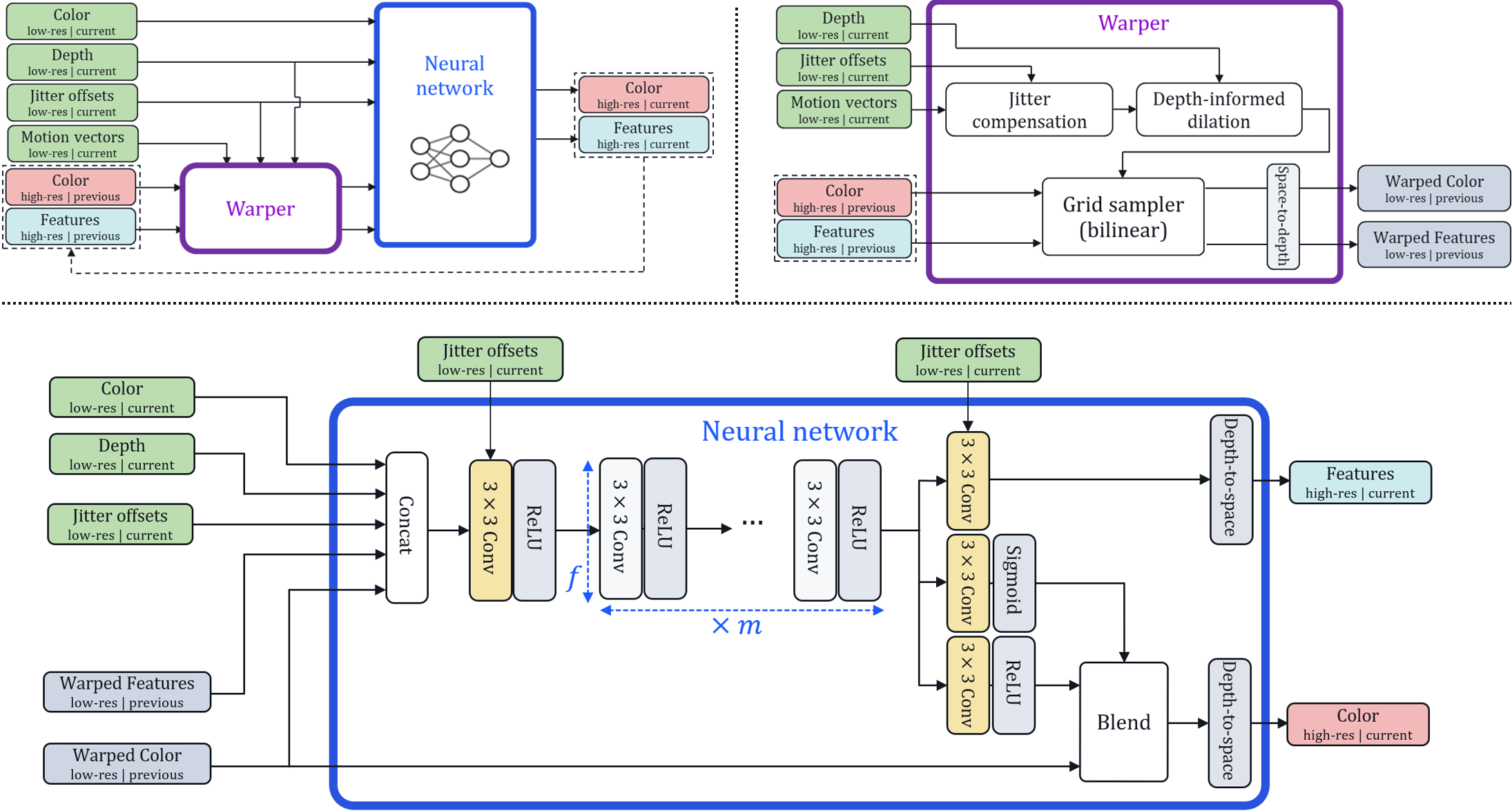}
\caption{High-level overview of our multi-frame supersamping approach \textit{(top-left)} and detailed description of its individual components: a warping module \textit{(top-right)} and a reconstruction neural network \textit{(bottom)}. $m$ and $f$ refer to the number of intermediate conv layers and the number of features in these layers, respectively.}
\label{figure:networkArchitecture}
\end{center}
\vspace{-0.7cm}
\end{figure*}

\vspace{-0.1cm}
\paragraph{Gaming supersampling.}

Temporal Anti-Aliasing (TAA) \cite{yang2009amortized, yang2020survey, Karis2014} and its upscaling forms \cite{herzog2010spatio, games2018unreal, fsr2_2022} exploit samples from past frames to recover missing details in the current frame. Compared to single-frame anti-aliasing techniques \cite{akeley1993reality, fxaa2009, reshetov2009morphological}, TAA has gained popularity over the past decade as it provides a good trade-off between accuracy and practicality, even in the context of deferred rendering where MSAA \cite{akeley1993reality} becomes bandwidth prohibitive. A typical TAA pipeline includes a re-projection step \cite{nehab2007accelerating, scherzer2007pixel} to re-align historical data using accurate motion vectors, a history validation step to reject or rectify past samples that are invalid or stale due to \eg occlusion, lighting or shading changes, and a blending (or accumulation) step to produce the final output. While TAA-based super-resolution approaches such as FSR2 \cite{fsr2_2022} leverage hand-engineered heuristics 
to perform history validation and accumulation, DLSS \cite{liu2020dlss}, XeSS \cite{chowdhury2022intel} and Xiao \etal's work \cite{xiao2020neural} have showed that these steps can be replaced by a neural net. In the rest of the paper, we compare our algorithm mainly against Xiao \etal's \cite{xiao2020neural}, as the implementation details of DLSS and XeSS are not publicly available and, therefore, not reproducible.

\vspace{-0.1cm}
\paragraph{Graphics features traditonally used with supersampling.} Viewport jittering and negative mipmap biasing are two rendering techniques that are traditionally used to boost super-resolution accuracy. Viewport jittering consists in applying a sub-pixel shift to the camera sampling grid, and it is most useful when the camera is stationary as it ensures that consecutive frames contain complementary information about the scene. The subpixel offset typically follows a fixed-length sequence parameterized by, for example, a Halton sequence. Negative mipmap biasing, on the other hand, reduces the amount of low-pass prefiltering applied to textures, resulting in low-resolution renders with more high-frequency details.


\vspace{-0.2cm}
\paragraph{Related datasets} While there are many datasets for single-frame \cite{div2k, set5, set14, bsd100, urban100} and video \cite{Nah_2019_CVPR_Workshops_REDS, xue2019video} super-resolution of natural content, there is no publicly available dataset for gaming super-resolution. To the best of our knowledge, among existing datasets, Sintel \cite{Butler:ECCV:2012} would be the closest candidate as it consists of synthetic images and provides motion vectors. It is however available at only one resolution, which is problematic because DL-based super-resolution models trained to reconstruct images from artificially downsized images tend to overfit the degradation method \cite{liu2022blind}. Besides, Sintel does not provide jittered or mipbiaised samples, two key ingredients for gaming supersampling. Our dataset is also significantly larger than Sintel ($5\times$ more frames, and available at higher resolution).

\begin{figure*}[t!]
\begin{center}
   \includegraphics[width=\linewidth]{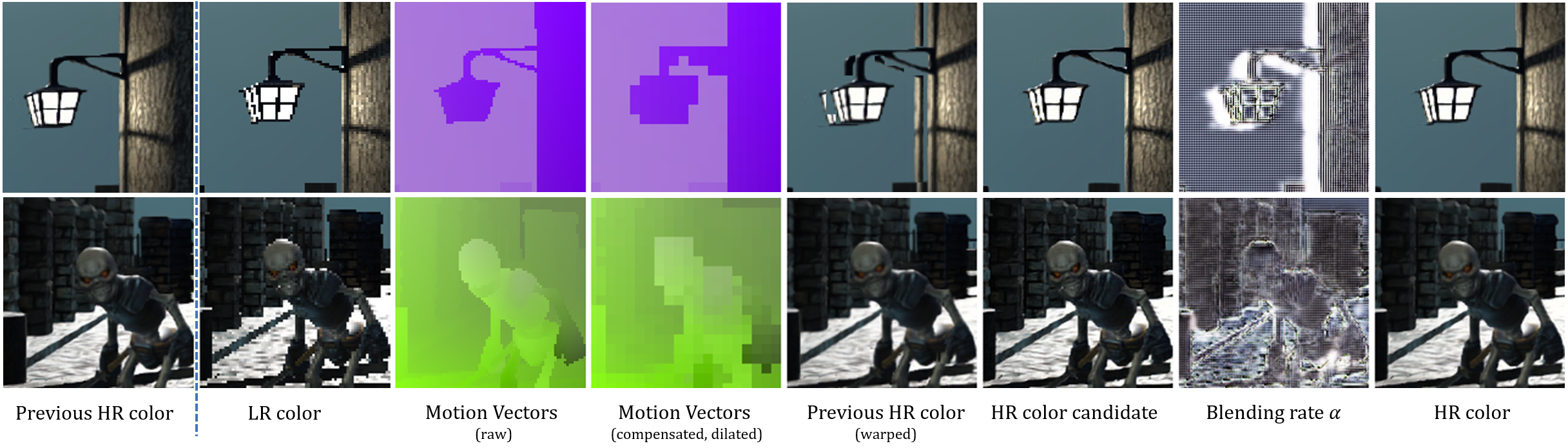}
\end{center}
   \caption{Visualization of data instances used at different steps of the algorithm \textit{(from left to right)}: the previous frame's high-resolution output, the current low-resolution render, the motion vectors before and after pre-processing, the re-projected output from the previous frame, the new high-resolution color candidate for the current frame, the blending mask $\alpha$, and the final output for the current frame. Note that a low value of $\alpha$ (dark) means that the color from the previous output is retained; a high value of $\alpha$ (bright) means that the color from the previous timestep is discarded in favour of the new candidate color.}
   \label{figure:networkIntermediateCrop}
\end{figure*}

\begin{figure}[ht!]
\begin{center}
   \includegraphics[width=\linewidth]{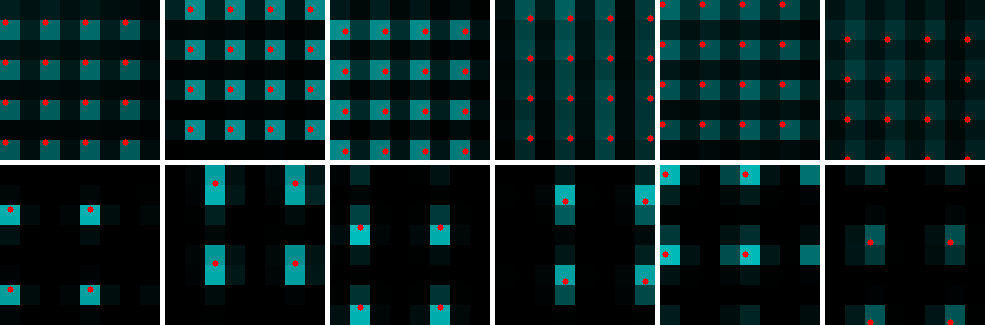}
\end{center}
   \caption{Visualization of the blending mask ($8\!\times\!8$ crops) over consecutive timesteps along with the corresponding subpixel-shifted sampling locations (red dots) on a surface that does not need history rejection for $2\times$ (first row) and $4\times$ upscaling (second row). When the previous samples are still relevant for the current frame, the model tends to only update the pixels at the sampling location.}
   \label{fig:maskjitter} 
\vspace{-0.2cm}
\end{figure}

\section{The Qualcomm Rasterized Images for Super-resolution Processing dataset}

The \emph{Qualcomm Rasterized Images for Super-resolution Processing} dataset, referred later in this paper as QRISP, was specifically designed to facilitate the development and research of super-resolution algorithms for gaming applications. To the best of our knowledge, this dataset has no publicly available equivalent.

\vspace{-0.2cm}
\paragraph{Data modalities.} The QRISP dataset consists of sequences of rasterized images captured at 60 frames per second with multiple modalities rendered at different resolutions ranging from 270p to 1080p.  For each frame, the dataset includes color, depth, and motion vectors with different properties such as mipmapbiasing, jittering, or both. Jittered samples are achieved by shifting the camera using a sub-pixel offset drawn from a cyclic Halton$(2, 3)$ sequence of length 16 and we occasionally include stationary segments in the camera path to make models trained on this dataset more robust to the ``static'' scenario. For low-resolution renders, we disable MSAA or any other frame-blurring anti-aliasing techniques and adjust the texture mip levels using a negative offset set to $-log_{2}(S)$ where $S$ is the per-dimension scaling factor, as typically done in gaming supersampling \cite{Karis2014, liu2020dlss, fsr1, fsr2_2022, xiao2020neural}. For high-res images, we target high-quality 1080p color images which were obtained by 2x-downsizing 2160p renders with MSAAx8 applied, as done in \cite{xiao2020neural}. Figure \ref{fig:modalities} shows an example of such an ``enhanced'' target image (see the bottom-right crop), along with the corresponding low-resolution renders.

\vspace{-0.2cm}
\paragraph{Scene diversity and composition.} The dataset is diverse, with a variety of backgrounds and models to enable better generalization to new video games. There are 13 scenes in total, with 10 scenes allocated for training and the remaining 3 reserved for evaluation. Some of these scenes can be seen in Figure \ref{fig:dataset} and more samples can be found in the supplementary material. The data was generated using the Unity game engine \cite{haas2014history}, with 3D assets sourced either from the Unity Asset Store \footnote{https://assetstore.unity.com/}, or from open-source projects. The list of Unity assets used in this work can be found in the supplementary material. To make the data more representative of realistic gaming scenarios, animated characters were incorporated to the scene. We also added textual UI elements on top of animated characters to make the algorithms more robust to elements without associated depth or motion vector information. 

\paragraph{Commercial baselines.} In this dataset, we have also included images upscaled by commercial solutions integrated into Unity on the same frames used for evaluation. At the time of the dataset collection, these included Nvidia's DLSS 2.2 and AMD's FSR 1.2\footnote{We do not compare against FSR 1.2 in this paper as we focus on multi-frame supersampling approaches.}, which can serve as reference baselines to assess the performance of new algorithms.

 We believe that releasing this dataset will be beneficial for many, as it provides a time-saving alternative to extracting synchronized LR-HR pairs from a game engine, with additional modalities such as depth and motion vectors, and properties like jittering or mipmap biaising. This dataset was primarily created to advance the development of super-resolution algorithms for gaming applications, but we believe that it can also be useful for other tasks, such as optical flow estimation.








\begin{figure*}[t!]
\begin{center}
   \includegraphics[width=\linewidth]{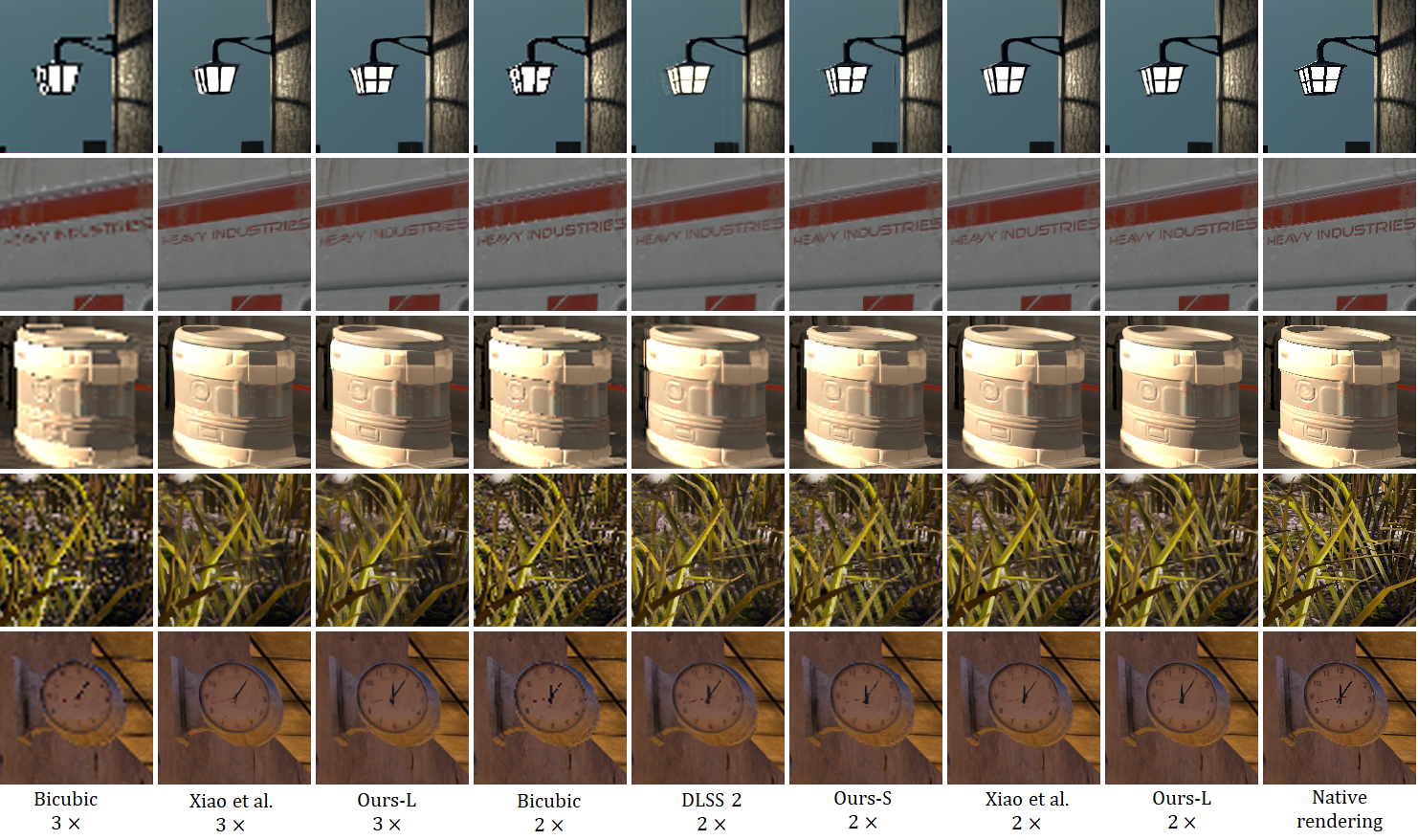}
\end{center}
   \caption{Super-resolution results ($2\times$ and $3\times$) by our algorithm vs bicubic upscaling, Xia \etal's approach \cite{xiao2020neural} and DLSS 2.2 \cite{liu2020dlss}.}
   \label{figure:results}
\end{figure*}

\begin{table*}[]
\small
\begin{center}
\begin{tabular}{cl|c|ccc|ccc|l|ccc|}
\cline{3-9} \cline{11-13}
\multicolumn{1}{l}{}                 &           & \multicolumn{1}{l|}{\textbf{Ours-S}} & \multicolumn{3}{c|}{\textbf{Ours-M}}          & \multicolumn{3}{c|}{\textbf{Ours-L}}          &           & \multicolumn{3}{c|}{\textbf{Xiao \etal}} \\ \cline{1-1} \cline{3-9} \cline{11-13} 
\multicolumn{1}{|c|}{Scaling factor} &           & $2\times$                            & $2\times$     & $3\times$     & $4\times$     & $2\times$     & $3\times$     & $4\times$     &           & $2\times$         & $3\times$        & $4\times$        \\ \cline{1-1} \cline{3-9} \cline{11-13} 
\multicolumn{1}{|c|}{MV dilation}    &           & 0.21                                 & 0.21          & 0.21          & 0.21          & 0.21          & 0.21          & 0.21          &           & -                 & -                & -                \\ \cline{1-1} \cline{3-9} \cline{11-13} 
\multicolumn{1}{|c|}{Warping}        &           & 0.15                                 & 0.15          & 0.15          & 0.15          & 0.15          & 0.15          & 0.15          &           & 0.92              & 0.92             & 0.92             \\ \cline{1-1} \cline{3-9} \cline{11-13} 
\multicolumn{1}{|c|}{Neural network} &           & 0.72                                 & 1.17          & 0.81          & 0.70          & 3.16          & 1.74          & 1.26          &           & 13.01             & 12.92            & 12.88            \\ \cline{1-1} \cline{3-9} \cline{11-13} 
\multicolumn{1}{|c|}{\textbf{Total}} & \textbf{} & \textbf{1.08}                        & \textbf{1.53} & \textbf{1.17} & \textbf{1.06} & \textbf{3.52} & \textbf{2.10} & \textbf{1.62} & \textbf{} & \textbf{13.93}    & \textbf{13.84}   & \textbf{13.80}   \\ \cline{1-1} \cline{3-9} \cline{11-13} 
\end{tabular}
\end{center}
\caption{Profiling results (in milliseconds) of our $2\times$, $3\times$ or $4\times$ architectures vs Xiao \etal on an RTX 3090 at 1080p target resolution. These timings were all obtained using Nvidia's TensorRT  \cite{tensorRT} for the neural network execution, using FP16 precision.} 
\label{tab:timings}
\vspace{-0.3cm}
\end{table*}


\section{Proposed algorithm}

Our proposed neural algorithm for gaming super-sampling consists of two components: a warping module for re-projecting historical data to the current frame, and a neural network for reconstructing the current image at target resolution. The reconstruction neural network blends a candidate image with the output image from the previous timestep using subpixel accurate motion vectors for motion compensation. Figures \ref{figure:networkArchitecture} and \ref{figure:networkIntermediateCrop} provide an overview of the solution and a visualization of data instances at various steps of the algorithm, respectively.


\subsection{Warping module}

As seen in the top-right diagram of Figure \ref{figure:networkArchitecture}, the motion compensation in our proposed algorithm is divided into three steps, which are described in detail below.

\paragraph{Jitter compensation.} We remove the viewport jittering contribution to the motion vector values. This is achieved by adding the jitter offset at frame $t-1$ and substracting the jitter offset at frame $t$ from the motion vector at frame $t$:

\begin{equation}
    MV_t = MV_t + J_{t-1} - J_t
\end{equation}

\paragraph{Depth-informed dilation.} This step modifies the motion vectors to reduce aliasing of foreground objects in re-projected images. This is achieved by producing a high-resolution block-based motion vector grid, where each block contains the motion vector value of the frontmost (i.e. lowest depth value) pixel within the block. In our experiments, we use a block size of $8\!\times\!8$ at high resolution. Similar ideas have been used in \cite{Karis2014, yang2020survey, fsr2_2022}.





\paragraph{Re-projection.} The preprocessed motion vectors are used to perform a bilinear warp and realign the previous timestep's high-resolution color images and neural features to the current frame. A space-to-depth operation is then applied to map the warping outputs to the input resolution, as in FRVSR \cite{sajjadi2018frame}. 

\subsection{Neural network}

Our neural network architecture is similar to the efficient single-frame super-resolution architectures of \cite{abpn, quicksrnet}. We use 3x3 Conv-ReLU blocks and a relatively small number of layers and channels. The output is mapped to high resolution using a depth-to-space operation. We however modify the architecture to make it suitable for multi-frame inputs and jittering:

\paragraph{Additional inputs and outputs.} In addition to color information \(C_t \), the neural network $F$ takes as inputs depth \(D_t \),  the jitter offset \(J_t \), the previous color output \(Y_{t-1} \) and features \(f_{t-1} \) re-aligned by the warper \(W\). It returns a pixel-wise blending mask $\alpha$ obtained using a sigmoid activation, a high-resolution candidate color image $\widetilde{Y{t}}$, and recurrent features \(f_{t} \) for the next timestep:

\begin{equation}
\alpha, \widetilde{Y_{t}}, f_{t} = F(C_t, D_t, J_t, W(Y_{t-1}), W(f_{t-1})) 
\end{equation}

\paragraph{Blending.} The candidate color image returned by the neural net is combined with the previous output using \( \alpha \):

\begin{equation}
 Y_{t} = \alpha * \widetilde{Y_{t}} + (1 - \alpha) * W(Y_{t-1})
\end{equation}

Figure \ref{fig:maskjitter} illustrates how the blending mask evolves over consecutive timesteps and shows that the model tends to use the candidate pixels located at the current sampling location but retains the previous samples everywhere else ($\alpha \approx 0$). The neural net is also able to identify and discard samples ($\alpha \approx 1$) from the re-projected color image that are outdated due to appearance changes or dis-occlusion (see Figure \ref{figure:networkIntermediateCrop}).


\paragraph{Jitter-conditioned convolutions.} To facilitate alignment of the low-resolution color input, which is sub-pixel shifted due to jittering, we predict the kernel weights of the first and last convolution modules using an MLP conditioned on the jitter offset \(J_t \). This is different from kernel prediction networks commonly used in denoising tasks \cite{thomas2022temporally, zhou2019spatio, mildenhall2018burst} or burst image super-resolution \cite{wronski2019handheld,cho2021weighted} where a separate kernel is predicted for each pixel, as we only predict one kernel for the entire frame, using a two-dimensional vector (i.e. the jitter offset for the current frame) as the only conditioning variable. At inference time, the jittering sequence is known in advance, so we pre-compute the kernel weights for each jitter offset and re-load the corresponding kernel whenever a new jittered frame is generated. This allows the neural network to more accurately realign the subpixel-shifted input data to the target resolution grid, with no additional computational overhead. The concurrent work of \cite{herveau2023minimal} used a similar technique for temporal anti-aliasing.

\paragraph{Comparison to previous work.} Xiao \etal's neural architecture consists of a feature extraction network that processes the current low-resolution frame, an upsampling module, a warping module that recursively realigns a rolling buffer of four upsampled feature maps, a feature reweighting network for history rejection, and a U-Net-style \cite{ronneberger2015u} reconstruction network to produce the final output. In comparison, our approach better leverages viewport jittering and has notable advantages in terms of speed and memory. First, all convolutional layers in our architecture run at the input resolution. Second, our historical data only consists of $4$ high-resolution channels compared to $48$ for Xiao \etal, resulting in a larger memory footprint and a higher latency for the re-alignment step.





\section{Implementation details}

 We experiment with three variants of our models: \textit{Ours-S (f16-m1)}, \textit{Ours-M (f32-m3)} and \textit{Ours-L (f64-m5)}, where \textit{f$X$-m$Y$} means that the architecture has $Y$ intermediate conv layers and $X$ feature channels. We adjust the number of recurrent features produced at low-resolution based on the scaling factor to end up with a single channel of features at high-resolution after depth-to-space. To predict the kernels of the first and last convs, we use a 7-layer MLP with 2048 hidden features and ReLU activations. Since our dataset uses the same fixed jittering sequence for the entire dataset, we also tried  optimizing a set of 16 kernels, but it resulted in worse performance (0.1 dB PSNR drop on test scenes).

We use mini-batches of eight 16-frame clips with spatial resolution $264\times264$ (at high resolution), an L1 loss, and train for 500k iterations using the Adam optimizer \cite{adam} with an initial learning rate of $1e-4$, decaying the learning rate by a factor 2 after 200k and 400k iterations. We optimize the models on $80\%$ of the segments from each training scene and use the rest for validation.  



\begin{table}[t]
\small
\begin{center}
\begin{tabular}{ccccc}
\hline
\multicolumn{1}{|c|}{\multirow{2}{*}{Upscaling}} & \multicolumn{1}{c|}{\multirow{2}{*}{Method}}   & \multicolumn{3}{c|}{Metric}                                                                                      \\ \cline{3-5} 
\multicolumn{1}{|c|}{}                                  & \multicolumn{1}{c|}{}                          & \multicolumn{1}{c|}{PSNR}           & \multicolumn{1}{c|}{SSIM}            & \multicolumn{1}{c|}{LPIPS}          \\ \hline
                                                        &                                                &                                     &                                      &                                     \\ \hline
\multicolumn{1}{|c|}{\multirow{6}{*}{$2\times$}}               & \multicolumn{1}{c|}{Bicubic}                   & \multicolumn{1}{c|}{29.51}          & \multicolumn{1}{c|}{0.8672}          & \multicolumn{1}{c|}{0.219}          \\ \cline{2-5} 
\multicolumn{1}{|c|}{}                                  & \multicolumn{1}{c|}{DLSS 2}                    & \multicolumn{1}{c|}{30.21}          & \multicolumn{1}{c|}{0.8816}          & \multicolumn{1}{c|}{0.187}          \\ \cline{2-5} 
\multicolumn{1}{|c|}{}                                  & \multicolumn{1}{c|}{Xiao \etal} & \multicolumn{1}{c|}{31.89}          & \multicolumn{1}{c|}{0.9075}          & \multicolumn{1}{c|}{0.140}          \\ \cline{2-5} 
\multicolumn{1}{|c|}{}                                  & \multicolumn{1}{c|}{Ours-S (f16-m1)}           & \multicolumn{1}{c|}{31.18}          & \multicolumn{1}{c|}{0.8941}          & \multicolumn{1}{c|}{0.160}          \\ \cline{2-5} 
\multicolumn{1}{|c|}{}                                  & \multicolumn{1}{c|}{Ours-M (f32-m3)}           & \multicolumn{1}{c|}{31.80}          & \multicolumn{1}{c|}{0.9044}          & \multicolumn{1}{c|}{0.140}          \\ \cline{2-5} 
\multicolumn{1}{|c|}{}                                  & \multicolumn{1}{c|}{Ours-L (f64-m5)}           & \multicolumn{1}{c|}{\textbf{32.21}} & \multicolumn{1}{c|}{\textbf{0.9115}} & \multicolumn{1}{c|}{\textbf{0.134}} \\ \hline
                                                        &                                                &                                     &                                      &                                     \\ \hline
\multicolumn{1}{|c|}{\multirow{4}{*}{$3\times$}}               & \multicolumn{1}{c|}{Bicubic}                   & \multicolumn{1}{c|}{27.61}          & \multicolumn{1}{c|}{0.8034}          & \multicolumn{1}{c|}{0.322}          \\ \cline{2-5} 
\multicolumn{1}{|c|}{}                                  & \multicolumn{1}{c|}{Xiao \etal} & \multicolumn{1}{c|}{30.24}          & \multicolumn{1}{c|}{0.8729}          & \multicolumn{1}{c|}{0.200}          \\ \cline{2-5} 
\multicolumn{1}{|c|}{}                                  & \multicolumn{1}{c|}{Ours-M (f32-m3)}           & \multicolumn{1}{c|}{30.23}          & \multicolumn{1}{c|}{0.8655}          & \multicolumn{1}{c|}{0.203}          \\ \cline{2-5} 
\multicolumn{1}{|c|}{}                                  & \multicolumn{1}{c|}{Ours-L (f64-m5)}           & \multicolumn{1}{c|}{\textbf{30.67}} & \multicolumn{1}{c|}{\textbf{0.8747}} & \multicolumn{1}{c|}{\textbf{0.187}} \\ \hline
                                                        &                                                &                                     &                                      &                                     \\ \hline
\multicolumn{1}{|c|}{\multirow{4}{*}{$4\times$}}               & \multicolumn{1}{c|}{Bicubic}                   & \multicolumn{1}{c|}{26.42}          & \multicolumn{1}{c|}{0.7535}          & \multicolumn{1}{c|}{0.391}          \\ \cline{2-5} 
\multicolumn{1}{|c|}{}                                  & \multicolumn{1}{c|}{Xiao \etal} & \multicolumn{1}{c|}{29.02}          & \multicolumn{1}{c|}{0.8364}          & \multicolumn{1}{c|}{0.259}          \\ \cline{2-5} 
\multicolumn{1}{|c|}{}                                  & \multicolumn{1}{c|}{Ours-M (f32-m3)}           & \multicolumn{1}{c|}{29.06}          & \multicolumn{1}{c|}{0.8305}          & \multicolumn{1}{c|}{0.258}          \\ \cline{2-5} 
\multicolumn{1}{|c|}{}                                  & \multicolumn{1}{c|}{Ours-L (f64-m5)}           & \multicolumn{1}{c|}{\textbf{29.42}} & \multicolumn{1}{c|}{\textbf{0.8403}} & \multicolumn{1}{c|}{\textbf{0.238}} \\ \hline
\end{tabular}
\end{center}
\caption{PSNR, SSIM and LPIPS scores for our model, DLSS 2.2 \cite{liu2020dlss} and our implementation of Xiao \etal \cite{xiao2020neural} for $2\times$, $3\times$ and $4\times$ upscaling. For a more fine-grained analysis, a per-scene breakdown of PSNR and SSIM scores is available in the supplementary material.}
\label{table:results}
\vspace{-0.1cm}
\end{table}

\begin{table}[t]
\small
\begin{center}
\begin{tabular}{lclcc}
\hline
\multicolumn{1}{|c|}{\multirow{2}{*}{Model   variants}} & \multicolumn{2}{c|}{Entire test set}                             & \multicolumn{2}{c|}{Static frames}                         \\ \cline{2-5} 
\multicolumn{1}{|c|}{}                                  & \multicolumn{1}{c|}{PSNR}  & \multicolumn{1}{c|}{LPIPS}  & \multicolumn{1}{c|}{PSNR}   & \multicolumn{1}{c|}{Pixel Std} \\ \hline
                                                        & \multicolumn{1}{l}{}       &                             & \multicolumn{1}{l}{}        & \multicolumn{1}{l}{}           \\ \hline
\multicolumn{1}{|l|}{Baseline (f32-m3)}                 & \multicolumn{1}{c|}{31.80} & \multicolumn{1}{l|}{0.140} & \multicolumn{1}{c|}{37.38}  & \multicolumn{1}{c|}{0.54}      \\ \hline
                                                        & \multicolumn{1}{l}{}       &                             & \multicolumn{1}{l}{}        & \multicolumn{1}{l}{}           \\ \hline
\multicolumn{1}{|l|}{(-)  MV dilation}                  & \multicolumn{1}{c|}{31.97} & \multicolumn{1}{l|}{0.136} & \multicolumn{1}{c|}{37.47}  & \multicolumn{1}{c|}{0.54}      \\ \hline
\multicolumn{1}{|l|}{(-) blending}                      & \multicolumn{1}{c|}{31.80} & \multicolumn{1}{l|}{0.144} & \multicolumn{1}{c|}{36.86}  & \multicolumn{1}{c|}{1.10}      \\ \hline
\multicolumn{1}{|l|}{(-) jitter}                        & \multicolumn{1}{c|}{31.61} & \multicolumn{1}{l|}{0.153} & \multicolumn{1}{c|}{35.06}  & \multicolumn{1}{c|}{0.03}      \\ \hline
                                                        & \multicolumn{1}{l}{}       &                             & \multicolumn{1}{l}{}        & \multicolumn{1}{l}{}           \\ \hline
\multicolumn{1}{|l|}{(-) MVs, (+) RAFT}                 & \multicolumn{1}{c|}{31.09} & \multicolumn{1}{l|}{0.169} & \multicolumn{1}{c|}{35.77}  & \multicolumn{1}{c|}{1.21}      \\ \hline
\multicolumn{1}{|l|}{(-) warping}                       & \multicolumn{1}{c|}{30.69} & \multicolumn{1}{l|}{0.183} & \multicolumn{1}{c|}{37.35}  & \multicolumn{1}{c|}{0.53}      \\ \hline
\multicolumn{1}{|l|}{(-) multiple frames}               & \multicolumn{1}{c|}{30.56} & \multicolumn{1}{l|}{0.192} & \multicolumn{1}{c|}{33.48}  & \multicolumn{1}{c|}{1.96}      \\ \hline
                                                        & \multicolumn{1}{l}{}       &                             & \multicolumn{1}{l}{}        & \multicolumn{1}{l}{}           \\ \hline
\multicolumn{1}{|l|}{(-) first jitter conv}             & \multicolumn{1}{c|}{31.68} & \multicolumn{1}{l|}{0.145} & \multicolumn{1}{c|}{37.40}  & \multicolumn{1}{c|}{0.49}      \\ \hline
\multicolumn{1}{|l|}{(-) last jitter conv}              & \multicolumn{1}{c|}{31.50} & \multicolumn{1}{l|}{0.148} & \multicolumn{1}{c|}{37.38}  & \multicolumn{1}{c|}{0.56}      \\ \hline
\multicolumn{1}{|l|}{(-) jitter conv}                   & \multicolumn{1}{c|}{31.26} & \multicolumn{1}{l|}{0.156} & \multicolumn{1}{c|}{37.20}  & \multicolumn{1}{c|}{0.56}      \\ \hline
                                                        &                            &                             &                             &                                \\ \hline
\multicolumn{1}{|l|}{(-2) training scenes}              & \multicolumn{1}{c|}{31.74} & \multicolumn{1}{l|}{0.144} & \multicolumn{1}{c|}{37.42}  & \multicolumn{1}{c|}{0.57}      \\ \hline
\multicolumn{1}{|l|}{(-5) training scenes}              & \multicolumn{1}{c|}{31.78} & \multicolumn{1}{l|}{0.142} & \multicolumn{1}{c|}{37.14}  & \multicolumn{1}{c|}{0.76}      \\ \hline
\multicolumn{1}{|l|}{(-8) training scenes}              & \multicolumn{1}{c|}{31.56} & \multicolumn{1}{l|}{0.152} & \multicolumn{1}{c|}{37.11}  & \multicolumn{1}{c|}{0.91}      \\ \hline
                                                        & \multicolumn{1}{l}{}       &                             & \multicolumn{1}{l}{}        & \multicolumn{1}{l}{}           \\ \hline
\multicolumn{1}{|l|}{(+) perceptual loss}               & \multicolumn{1}{l|}{31.72} & \multicolumn{1}{l|}{0.125} & \multicolumn{1}{c|}{37.40} & \multicolumn{1}{c|}{0.54}     \\ \hline
\end{tabular}
\end{center}
\caption{Ablation study. We report the average PSNR and LPIPS scores on the entire test set, as well as PSNRs and average pixel-wise standard deviation on a static segment from the AbandonedSchool scene.}
\label{table:ablation}
\end{table}

\begin{figure}[t!]
\begin{center}
   \includegraphics[width=\linewidth]{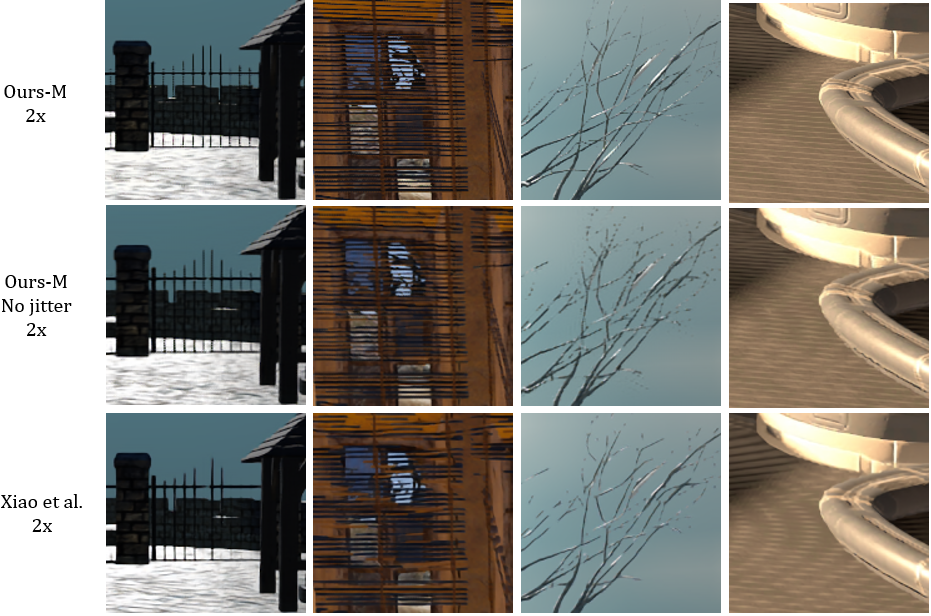}
\end{center}
   \caption{Visual impact of jittering on reconstructions of static scenes using our algorithm vs non-jittered alternatives.}
   \label{fig:jitter-importance}
\end{figure}

\begin{figure*}[t!]
\begin{center}
   \includegraphics[width=\linewidth]{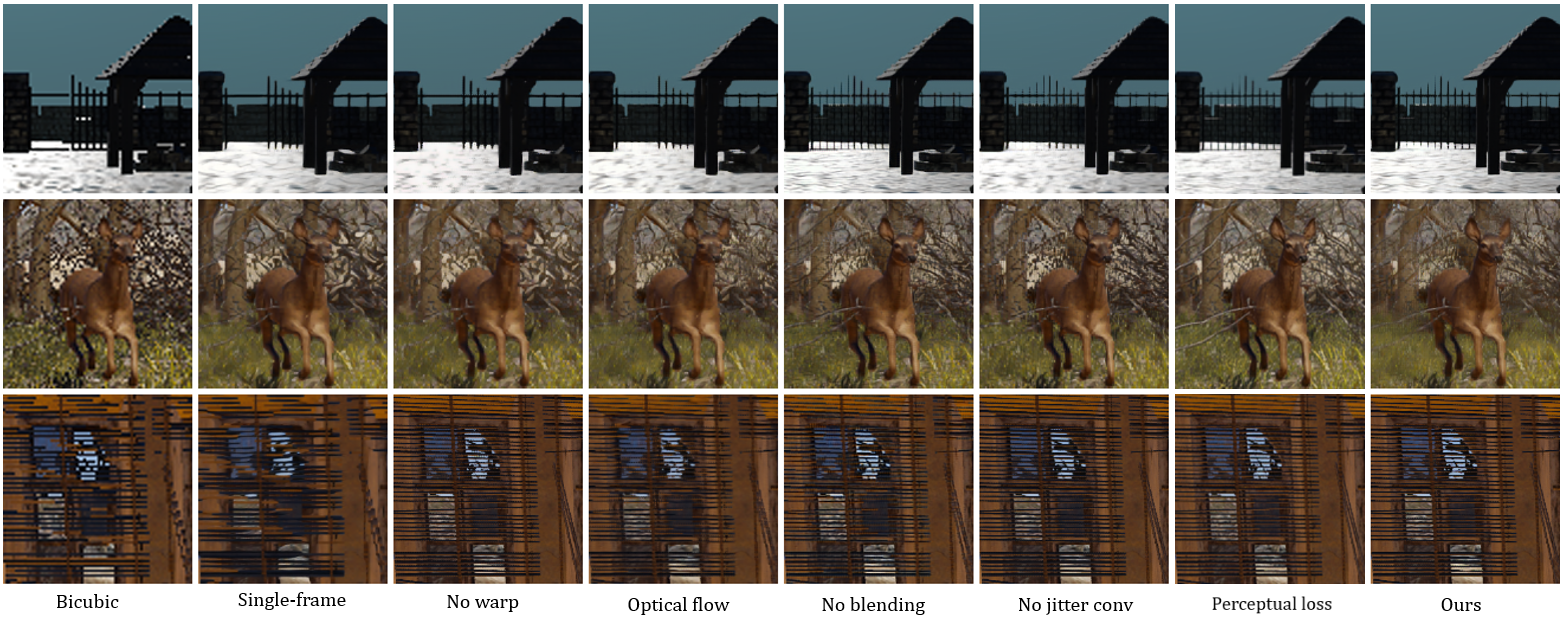}
\end{center}
   \caption{Ablation study using our $2\times$ medium-sized architecture. We ablate the following components \textit{(from left to right)}: all components except the reconstruction network which we run on a single low-resolution frame, warping, motion vectors (which we replace by estimated optical flow obtained using RAFT \cite{teed2020raft}), jitter-conditioned convolutions and the L1 loss which we replace to the loss defined in \cite{xiao2020neural}.}
   \label{figure:ablation}
\end{figure*}

\begin{figure}[t!]
\begin{center}
   \includegraphics[width=0.95\linewidth]{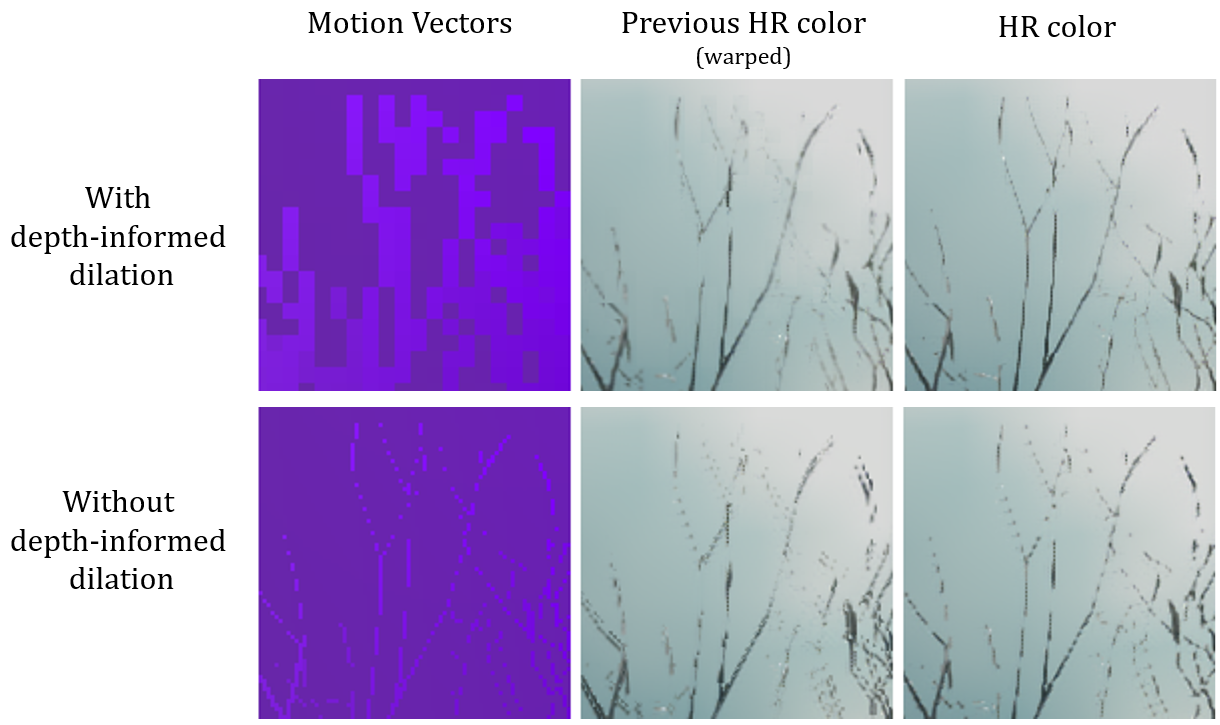}
\end{center}
   \caption{Reconstruction of thin objects, with and without depth-informed motion vector dilation.}
   \label{figure:dilation}
\end{figure}




\section{Results analysis}

\subsection{Speed-accuracy tradeoff compared to existing solutions} 

Table \ref{table:results} reports the average PSNR, SSIM and LPIPS scores obtained by our network, DLSS 2.2 and our implementation of Xiao \etal's solution \cite{xiao2020neural} on the test scenes. Our small model outperforms DLSS, while our large architecture is slightly better than Xiao \etal. These PSNR, SSIM and LPIPS improvements also manifest in visual quality improvements, as seen in Figure \ref{figure:results}. We observe that DLSS 2.2 produces more ghosting artifacts (visible on the lantern in the first row or on the barrel in the third row) than the other approaches.  The benefits from leveraging jittered samples are particularly visible in the reconstruction of static scenes (see Figure \ref{fig:jitter-importance}).

Table \ref{tab:timings} demonstrates that our larger model generates 1080p images in $3.52$ ms through $2\times$ upscaling on an Nvidia RTX 3090, representing a $4\times$ improvement compared to the architecture proposed by Xiao \etal\cite{xiao2020neural} while maintaining the same level of accuracy. Our small architecture runs in $1.08$ ms for the same workload.  Additionally, our architecture scales better to larger magnification factors, with our $4\times$ architecture offering an $8.5\times$ speedup compared to Xiao \etal\cite{xiao2020neural} for the same level of accuracy. We believe these timings could be improved using optimized CUDA kernels for the reprojection-related operations. 


\subsection{Ablation studies}

In this section, we ablate individual components from our $2\times$ medium-sized model, \textit{Ours-M}, to illustrate the impact of each component on visual quality and stability.

\paragraph{Depth-informed motion vector dilation.} While removing motion vector dilation improves PSNR and LPIPS scores (see Table \ref{table:ablation}), we found this step beneficial for the reconstruction of thin objects. This is visible in Figure \ref{figure:dilation} where we illustrate the effect of depth-informed dilation on the motion vectors, resulting in a warped image with less ghosting and aliasing artifacts, leading to a better reconstruction.

\paragraph{Reconstruction quality and temporal stability on static scenes.} We evaluate the quality and temporal stability of model outputs on a section of the AbandonedSchool scene\footnote{Segment 0001, from frame 275 to 292} where the camera is stationary. We report the average PSNR and pixel-wise standard deviation on these frames in Table \ref{table:ablation}, in addition to the average PSNR on the entire test set. We observe that:  

\begin{itemize}
    \item The single-frame variant of our architecture poorly reconstructs fine-grained details and is not temporally stable.

    \item Jittering is key to properly reconstruct static scenes: without it, the average PSNR on static scenes drops significantly. The benefits from leveraging jittered samples is also visible in Figure \ref{fig:jitter-importance}. 
    
    \item Blending improves temporal stability. Without it, the pixel-wise standard deviation doubles (from 0.54 to 1.10) and we generally observe considerably more flickering artifacts with this variant. 

    \item Temporal stability benefits from more (targeted) training data. The pixel-wise standard deviation score drops significant when the last 5 scenes are not used because only those contain static segments.
\end{itemize}


\paragraph{On the benefits of jitter-conditioned convolutions} Table \ref{table:ablation} shows the benefits of using jitter-conditioned kernels in the first and last convolution modules. Without these, we observe a 0.54 dB PSNR drop. Table \ref{table:ablation} also suggests that the last layer is the one that benefits the most from using a jitter-conditioned kernel.

\paragraph{On the importance of accurate motion compensation}

To quantify the importance of accurate motion compensation, we trained our architecture both without motion compensation and on top of estimated motion vectors, which were estimated using RAFT \cite{teed2020raft} with weights pre-trained on Sintel \cite{Butler:ECCV:2012}. In both cases, the results show a significant PSNR drop. When the camera remains static, the variant without motion compensation works well in terms of reconstruction and temporal stability (as seen in Table \ref{table:ablation}), but the quality drops to the level of a single frame architecture when the camera moves (see the first two rows of Figure \ref{figure:ablation}). 

\paragraph{Replacing the L1 loss to a perceptual loss} We find that the loss used in \cite{xiao2020neural} (\ie a weighted combination of SSIM and a VGG-based perceptual loss) improves the sharpness and perceived quality of high-frequency textures (most visible in the ``deer" crop from Figure \ref{figure:ablation}) at the cost of slightly more temporal inconsistencies. Quantitatively, this loss improves LPIPS scores significantly (from $0.140$ to $0.125$ for Ours-M) with minor PSNR and SSIM differences.


\section{Conclusion and future work}

In this work, we present a novel neural supersampling approach that is $4\times$ more efficient than previous published work by Xiao \etal \cite{xiao2020neural} while maintaining the same level of accuracy. We propose a new dataset, QRISP, specifically designed for the research and development of super-resolution algorithms for gaming applications. When trained on the proposed dataset, our algorithm outperforms DLSS 2.2 \cite{liu2020dlss} in terms of visual quality. We believe that QRISP fills a gap in the dataset landscape and can serve as a valuable resource to advance the state-of-the-art in super-resolution techniques for gaming content. In future work, we plan to investigate quantizing the neural network component of our approach as this has the potential to make the algorithm even more efficient \cite{nagel2021white, xlsr, quicksrnet, sr4bit}. We also plan to explore whether more sophisticated losses can address the blurriness that sometimes arises in the reconstruction of severely under-sampled high-frequency textures (\eg the grass and trees behind the deer in Figure \ref{figure:ablation}). 

{\small
\bibliographystyle{ieee_fullname}
\bibliography{egbib}
}

\clearpage


\setcounter{section}{0}
\renewcommand*{\thesection}{Annex \Alph{section}}
\renewcommand*{\thesubsection}{\Alph{section}.\arabic{subsection}}

\section{The QRISP dataset}

\begin{figure}[h!]
\begin{center}
   \includegraphics[width=\linewidth]{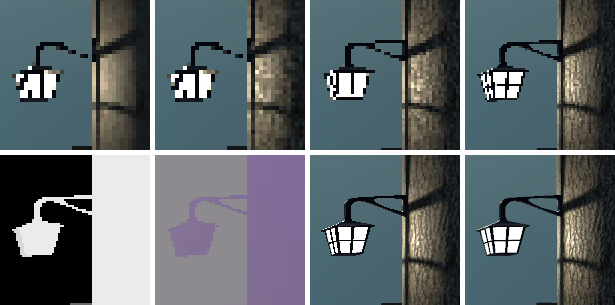}
\end{center}
   \caption{Example of data modalities available in the QRISP dataset. \textit{(first row, from left to right)}: Native 270p, Negative 2 mipmap biased 270p, Negative 1.58 mipmap biased 360p, Negative 1 mipmap biased 540p. \textit{(second row, from left to right):} 540p depth, 540p motion vectors, Native 1080p, Enhanced 1080p}
   \label{figure:appendix1}
\end{figure}

\subsection{Motivation}

The QRISP dataset was created to facilitate the development and research of super-resolution algorithms for gaming. The dataset consists of parallel captures of various scenes in different modalities and resolutions. It is designed to be diverse, with a variety of backgrounds and models, to better generalize to new video games. 

\subsection{Dataset modalities}

\begin{figure}[h!]

\begin{center}
   \includegraphics[width=\linewidth]{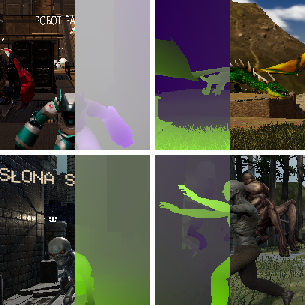}
\end{center}
   \caption{Example of animated characters and textual UI elements added manually to each scene. For each crop, we show both color and motion information to illustrate that these objects are moving and that the added UI does not have associated motion (nor depth) information.}
   \label{figure:appendix2}
\end{figure}

\begin{figure*}[h!]
\begin{center}
   \includegraphics[width=\linewidth]{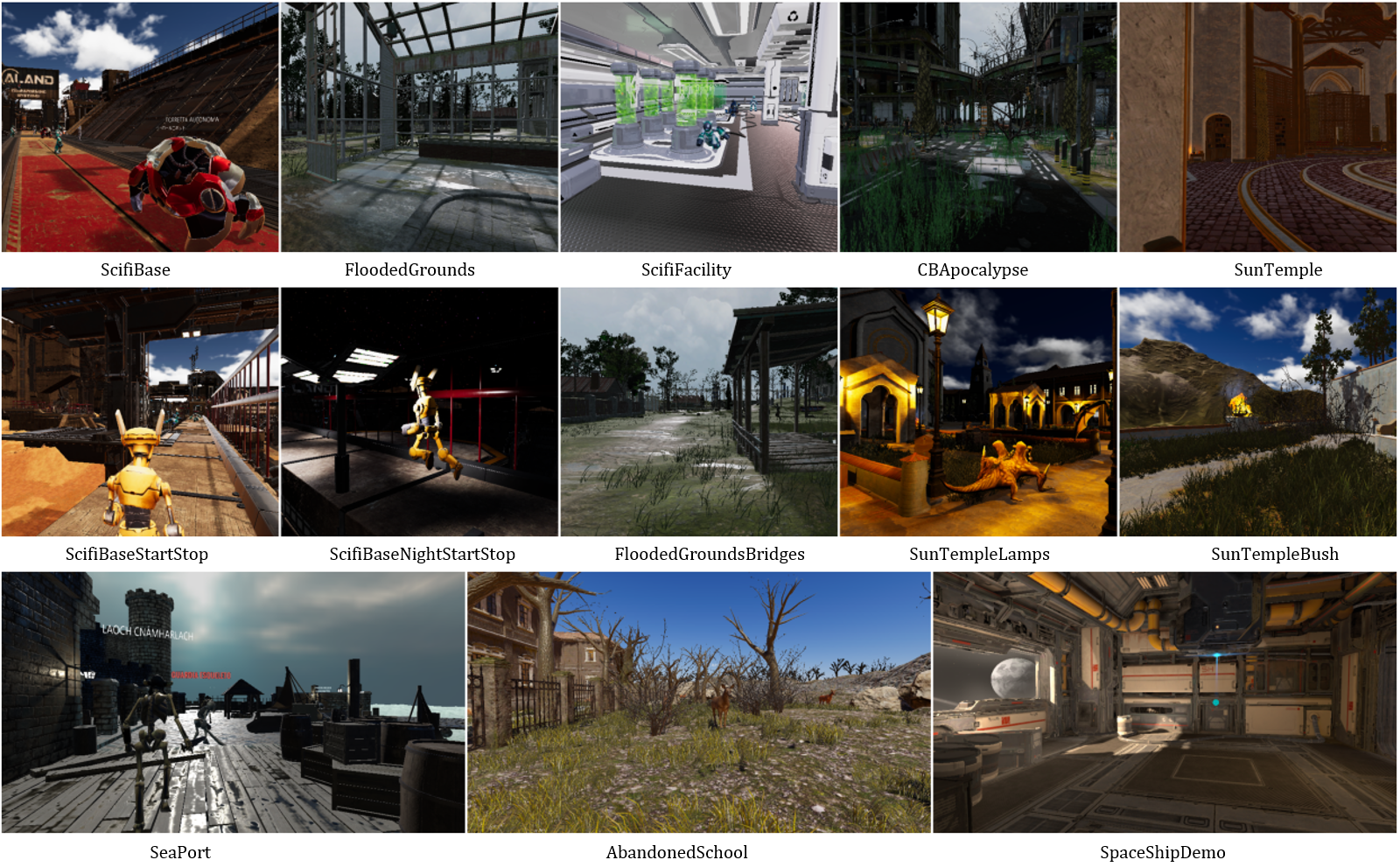}
\end{center}
   \caption{Overview of the 13 scenes from the QRISP dataset. The first two rows correspond to the 10 training scenes. The last rows shows the 3 scenes used for evaluation.}
   \label{figure:scenes}
\end{figure*}

The dataset consists of sequences of consecutive frames captured at 60 frames per second. For each frame, multiple modalities are rendered at different resolutions ranging from 270p to 1080p. Table \ref{table:modalities} provides a list of these modalities, which can be either generated using default parameters or mipmap biased, jittered, or both mipmap biased and jittered.  In addition to the modalities listed in the table, a JSON file containing camera parameters (including jitter offsets) is included for each segment. The following provides details about each type of modality:

\paragraph{Color} Rendered images are stored in 8-bit RGBA PNG files. In addition to low-resolution and native renders at 1080p resolution, we generate an additional color modality referred to as ``Enhanced''. This was produced by rendering color images at 2160p with MSAAx8 applied followed by 2x downsizing, resulting in a high-quality 1080p target image. 

\paragraph{Motion} Motion vectors are stored in a 16-bit EXR file, with the vertical velocity stored in the first channel and the horizontal velocity stored in the second. Unity uses a Y-up coordinate system so the vertical velocities may need to be negated to match the coordinate system used during motion compensation. Values are stored in the $[-1, 1]$ range, so they need to be scaled by the number of pixels in the corresponding dimension to convert the velocity in pixel units.

\paragraph{Depth} We save the depth information from the 32-bits z-buffer in the four channels of an 8-bit PNG file. The depth value corresponds to the distance between the rendered pixel and the camera near plane, scaled to the $[0, 1]$ range, where $0$ represents the near plane and 1 the far plane. To convert back from \([0, 255] ^ 4 \) to $[0, 1]$, we used the following equation:

\begin{equation}
depth = R / 255 + G / 255^2 + B / 255 ^ 3 + A / 255 ^ 4
\end{equation}

Where R, G, B, and A are the four 8-bit channels from the PNG file.

\paragraph{Jittering and other camera-related information}
For each segment, we provide a JSON file with camera intrinsic parameters (near plane, far plane and FOV) and frame-level information (jitter offset, camera position and orientation).


\subsection{Dataset composition and collection process}

The dataset consists of renders from 13 scenes in total, with 10 of them allocated for training and the remaining 3 reserved for evaluation.  Table \ref{table:scenes} provides a list of the scenes and the number of segments and frames available for each. 

\paragraph{Scenes compositions}
Our dataset includes 3D assets sourced either from the Unity Asset Store (\hyperlink{https://assetstore.unity.com/}{link}), or from open-source GitHub projects. To make the data more representative of real-world gaming scenarios, we manually add animated characters to the scene. We also occasionally add textual UI elements on top of animated characters to make the algorithms more robust to elements without associated depth or motion vector information. A list of the assets used is provided in Table \ref{table:scenesAssets}. 


\paragraph{Capture process}

In Unity, we use multiple ``twin'' cameras and shaders which all follow the same path to capture multiple resolutions and modalities simultaneously within the same frame. To ensure synchronization of pixels between the different renders, we may need to adjust certain parameters on a per-scene basis to make sure the rendering and animation is framerate dependent, like removing any asynchronous effects (e.g. wind physics for grass). We obtain jittered modalities by shifting the corresponding camera by a sub-pixel offset drawn from a cyclic Halton$(2, 3)$ sequence of length 16 and sometimes include ``stops'' in the camera path where the camera remains stationary.

\paragraph{Preprocessing/cleaning/labeling}

Frame-blurring post processing (bloom, motion blur, etc.) and anti-aliasing was disabled for all modalities, except for the ``Enhanced'' modality where we used MSAA as described earlier.  

\paragraph{Camera path and recording}

Sections of camera path were pre-recorded using the Unity game engine running at 60 fps (frames per second). During capturing, the camera followed the pre-recorded path through the scene and frames were captured at regular intervals of 30 frames for training data and 300 frames for evaluation data. 
In the case of training scenes including camera stops (see Table \ref{table:scenes}), the camera usually stayed stationary for 10 frames. Most training scenes have 120 frames between each segment of frames to increase diversity.

\subsection{Commercial baselines} 

In this dataset, we have also included images generated by commercial solutions integrated into Unity on the same frames used for evaluation. At the time of the dataset collection, these included Nvidia's DLSS 2.2.11.0 and AMD's FSR 1.2, which can serve as reference baselines to assess the performance of new algorithms.

\subsection{Potential use beyond super-resolution}

While this dataset was primarily created to facilitate the development of super-resolution algorithms for gaming applications; however, we believe that it could be useful for other tasks, such as optical flow estimation. 

\begin{table*}[t!]
\begin{center}
\begin{tabular}{|l|l|l|l|l|}
\hline
\textbf{Resolution}                                   & \textbf{Name}                                       & \multicolumn{1}{l|}{\textbf{Jittering applied?}} & \multicolumn{1}{l|}{\textbf{Mipmap bias offset}} & \textbf{Modality type} \\ \hline
\multicolumn{1}{|c|}{\multirow{2}{*}{1080p}} & Native                                     & No                                      & 0                                   & Color         \\ \cline{2-5} 
\multicolumn{1}{|c|}{}                       & Enhanced                                   & No                                      & NA                                  & Color         \\ \hline
\multicolumn{1}{|c|}{\multirow{7}{*}{540p}}  & DepthMipBiasMinus1                         & No                                      & -1                                  & Depth         \\ \cline{2-5} 
\multicolumn{1}{|c|}{}                       & DepthMipBiasMinus1Jittered                 & Yes                                     & -1                                  & Depth         \\ \cline{2-5} 
\multicolumn{1}{|c|}{}                       & MipBiasMinus1                         & No                                      & -1                                  & Color         \\ \cline{2-5} 
\multicolumn{1}{|c|}{}                       & MipBiasMinus1Jittered                 & Yes                                     & -1                                  & Color         \\ \cline{2-5} 
\multicolumn{1}{|c|}{}                       & MotionVectorsMipBiasMinus1            & No                                      & -1                                  & Motion vector \\ \cline{2-5} 
\multicolumn{1}{|c|}{}                       & MotionVectorsMipBiasMinus1Jittered    & Yes                                     & -1                                  & Motion vector \\ \cline{2-5} 
\multicolumn{1}{|c|}{}                       & Native                                & No                                      & 0                                   & Color         \\ \hline
\multicolumn{1}{|c|}{\multirow{7}{*}{360p}}                        & DepthMipBiasMinus1.58                 & No                                      & -1,58                               & Depth         \\ \cline{2-5} 
                                             & DepthMipBiasMinus1.58Jittered         & Yes                                     & -1,58                               & Depth         \\ \cline{2-5} 
                                             & MipBiasMinus1.58                      & No                                      & -1,58                               & Color         \\ \cline{2-5} 
                                             & MipBiasMinus1.58Jittered              & Yes                                     & -1,58                               & Color         \\ \cline{2-5} 
                                             & MotionVectorsMipBiasMinus1.58         & No                                      & -1,58                               & Motion vector \\ \cline{2-5} 
                                             & MotionVectorsMipBiasMinus1.58Jittered & Yes                                     & -1,58                               & Motion vector \\ \cline{2-5} 
                                             & Native                                & No                                      & 0                                   & Color         \\ \hline
\multicolumn{1}{|c|}{\multirow{7}{*}{270p}}                        & DepthMipBiasMinus2                    & No                                      & -2                                  & Depth         \\ \cline{2-5} 
                                             & DepthMipBiasMinus2Jittered            & Yes                                     & -2                                  & Depth         \\ \cline{2-5} 
                                             & MipBiasMinus2                         & No                                      & -2                                  & Color         \\ \cline{2-5} 
                                             & MipBiasMinus2Jittered                 & Yes                                     & -2                                  & Color         \\ \cline{2-5} 
                                             & MotionVectorsMipBiasMinus2            & No                                      & -2                                  & Motion vector \\ \cline{2-5} 
                                             & MotionVectorsMipBiasMinus2Jittered    & Yes                                     & -2                                  & Motion vector \\ \cline{2-5} 
                                             & Native                                & No                                      & 0                                   & Color         \\ \hline
\end{tabular}                    
\end{center}
\caption{Per-resolution breakdown of the modalities available in the QRISP dataset. Each modality can be either generated using default parameters, mipmap biased, jittered, or both mipmap biased and jittered.}
\label{table:modalities}
\end{table*}

\begin{table*}[t!]
\begin{center}
\begin{tabular}{llllll}
\hline
\multicolumn{1}{|l|}{\textbf{Scene}}          & \multicolumn{1}{l|}{\textbf{Split}} & \multicolumn{1}{l|}{\textbf{Segments}} & \multicolumn{1}{l|}{\textbf{Frames Per Segment}} & \multicolumn{1}{l|}{\textbf{Total Frames}} & \multicolumn{1}{l|}{\textbf{Includes stops?}} \\ \hline
\multicolumn{1}{|l|}{FloodedGrounds}          & \multicolumn{1}{l|}{Train}          & \multicolumn{1}{l|}{21}                & \multicolumn{1}{l|}{30}                          & \multicolumn{1}{l|}{630}                   & \multicolumn{1}{l|}{No}                       \\ \hline
\multicolumn{1}{|l|}{SciFifacility}           & \multicolumn{1}{l|}{Train}          & \multicolumn{1}{l|}{21}                & \multicolumn{1}{l|}{30}                          & \multicolumn{1}{l|}{630}                   & \multicolumn{1}{l|}{No}                       \\ \hline
\multicolumn{1}{|l|}{SunTemple}               & \multicolumn{1}{l|}{Train}          & \multicolumn{1}{l|}{36}                & \multicolumn{1}{l|}{30}                          & \multicolumn{1}{l|}{1080}                  & \multicolumn{1}{l|}{No}                       \\ \hline
\multicolumn{1}{|l|}{CB-Apocalypse}           & \multicolumn{1}{l|}{Train}          & \multicolumn{1}{l|}{30}                & \multicolumn{1}{l|}{30}                          & \multicolumn{1}{l|}{900}                   & \multicolumn{1}{l|}{No}                       \\ \hline
\multicolumn{1}{|l|}{SciFiBase}               & \multicolumn{1}{l|}{Train}          & \multicolumn{1}{l|}{41}                & \multicolumn{1}{l|}{30}                          & \multicolumn{1}{l|}{1230}                  & \multicolumn{1}{l|}{No}                       \\ \hline
\multicolumn{1}{|l|}{FloodedGroundsBridges}   & \multicolumn{1}{l|}{Train}          & \multicolumn{1}{l|}{20}                & \multicolumn{1}{l|}{30}                          & \multicolumn{1}{l|}{600}                   & \multicolumn{1}{l|}{Yes}                      \\ \hline
\multicolumn{1}{|l|}{ScifiBaseNightStartStop} & \multicolumn{1}{l|}{Train}          & \multicolumn{1}{l|}{20}                & \multicolumn{1}{l|}{30}                          & \multicolumn{1}{l|}{600}                   & \multicolumn{1}{l|}{Yes}                      \\ \hline
\multicolumn{1}{|l|}{ScifiBaseStartStop}      & \multicolumn{1}{l|}{Train}          & \multicolumn{1}{l|}{21}                & \multicolumn{1}{l|}{30}                          & \multicolumn{1}{l|}{630}                   & \multicolumn{1}{l|}{Yes}                      \\ \hline
\multicolumn{1}{|l|}{SunTempleBush}           & \multicolumn{1}{l|}{Train}          & \multicolumn{1}{l|}{11}                & \multicolumn{1}{l|}{30}                          & \multicolumn{1}{l|}{330}                   & \multicolumn{1}{l|}{Yes}                      \\ \hline
\multicolumn{1}{|l|}{SunTempleLamps}          & \multicolumn{1}{l|}{Train}          & \multicolumn{1}{l|}{21}                & \multicolumn{1}{l|}{30}                          & \multicolumn{1}{l|}{630}                   & \multicolumn{1}{l|}{Yes}                      \\ \hline
                                              &                                     &                                        &                                                  &                                            &                                               \\[-1ex] \hline
\multicolumn{1}{|l|}{AbandonedSchool}         & \multicolumn{1}{l|}{Test}           & \multicolumn{1}{l|}{2}                 & \multicolumn{1}{l|}{300}                         & \multicolumn{1}{l|}{600}                   & \multicolumn{1}{l|}{Yes}                      \\ \hline
\multicolumn{1}{|l|}{SpaceShipDemo}           & \multicolumn{1}{l|}{Test}           & \multicolumn{1}{l|}{2}                 & \multicolumn{1}{l|}{300}                         & \multicolumn{1}{l|}{600}                   & \multicolumn{1}{l|}{Yes}                      \\ \hline
\multicolumn{1}{|l|}{Seaport}                 & \multicolumn{1}{l|}{Test}           & \multicolumn{1}{l|}{1}                 & \multicolumn{1}{l|}{300}                         & \multicolumn{1}{l|}{300}                   & \multicolumn{1}{l|}{Yes}                      \\ \hline
                                              &                                     &                                        &                                                  &                                            &                                               \\[-1ex] \hline
\multicolumn{1}{|l|}{Total Training}          & \multicolumn{1}{l|}{-}               & \multicolumn{1}{l|}{242}               & \multicolumn{1}{l|}{30}                          & \multicolumn{1}{l|}{7260}                  & \multicolumn{1}{l|}{-}                         \\ \hline
\multicolumn{1}{|l|}{Total Test}              & \multicolumn{1}{l|}{-}               & \multicolumn{1}{l|}{5}                 & \multicolumn{1}{l|}{300}                         & \multicolumn{1}{l|}{1500}                  & \multicolumn{1}{l|}{-}                         \\ \hline
\multicolumn{1}{|l|}{Total}                   & \multicolumn{1}{l|}{-}               & \multicolumn{1}{l|}{-}                 & \multicolumn{1}{l|}{-}                           & \multicolumn{1}{l|}{8760}                  & \multicolumn{1}{l|}{-}                         \\ \hline
\end{tabular}
\end{center}
   \caption{List of dataset scenes, with split information, the number of segments and frames per segment, and whether it includes static segments where the camera remains stationary.}
   \label{table:scenes}
\end{table*}

\begin{table*}[h!]
\begin{center}
\begin{tabular}{|l|l|l|}
\hline
\textbf{Scene} & \textbf{Assets Used} & \textbf{Source} \\ \hline
\multirow{4}{*}{CB-Apocalypse} & CBU: Apocalypse Edition & Unity Asset Store \\ \cline{2-3} 
 & \begin{tabular}[c]{@{}l@{}}ULTIMATE ANIMATION \\ COLLECTION\end{tabular} & Unity Asset Store \\ \cline{2-3} 
 & Animal Pack Deluxe & Unity Asset Store \\ \cline{2-3} 
 & Customizable Survivors Pack & Unity Asset Store \\ \hline
\multirow{2}{*}{SciFifacility} & Sci-Fi Facility & Unity Asset Store \\ \cline{2-3} 
 & Robot Warriors Cartoon & Unity Asset Store \\ \hline
\multirow{4}{*}{\begin{tabular}[c]{@{}l@{}}FloodedGrounds\\ FloodedGroundsBridges\end{tabular}} & Flooded Grounds & Unity Asset Store \\ \cline{2-3} 
 & Ghoul-zombie & Unity Asset Store \\ \cline{2-3} 
 & Zombie & Unity Asset Store \\ \cline{2-3} 
 & Fantastic Creature \#1 & Unity Asset Store \\ \hline
\multirow{4}{*}{\begin{tabular}[c]{@{}l@{}}SciFiBase\\ ScifiBaseNightStartStop\\ ScifiBaseStartStop\end{tabular}} & Sci-Fi base & Unity Asset Store \\ \cline{2-3} 
 & Robot 1 & Unity Asset Store \\ \cline{2-3} 
 & Robot Warriors Cartoon & Unity Asset Store \\ \cline{2-3} 
 & Robot Sphere & Unity Asset Store \\ \hline
\multirow{3}{*}{SunTemple} & Sun Temple & Unity Asset Store \\ \cline{2-3} 
 & Animal Pack Deluxe & Unity Asset Store \\ \cline{2-3} 
 & Dragon for Boss Monster : HP & Unity Asset Store \\ \hline
\multirow{2}{*}{SunTempleBush} & Sun Temple & Unity Asset Store \\ \cline{2-3} 
 & Real Landscapes - Valley Forest & Unity Asset Store \\ \hline
\multirow{3}{*}{SunTempleLamps} & Sun Temple & Unity Asset Store \\ \cline{2-3} 
 & Dragon for Boss Monster : HP & Unity Asset Store \\ \cline{2-3} 
 & Flooded Grounds & Unity Asset Store \\ \hline
\multirow{2}{*}{AbandonedSchool} & Animal Pack Deluxe & Unity Asset Store \\ \cline{2-3} 
 & HQ Abandoned School (Modular) & Unity Asset Store \\ \hline
SpaceShipDemo & Space Ship Demo & \begin{tabular}[c]{@{}l@{}}https://github.com/Unity-\\ Technologies/SpaceshipDemo\end{tabular} \\ \hline
\multirow{3}{*}{Seaport} & Old Sea Port & Unity Asset Store \\ \cline{2-3} 
 & Fantasy Monster - Skeleton & Unity Asset Store \\ \cline{2-3} 
 & Dungeon Skeletons Demo & Unity Asset Store \\ \hline
\end{tabular}
\end{center}
   \caption{List of assets used for each scene.}
   \label{table:scenesAssets}
\end{table*}

\clearpage

\section{Supplementary results}



\begin{figure}[ht!]
\begin{center}
   \includegraphics[width=2.1\linewidth]{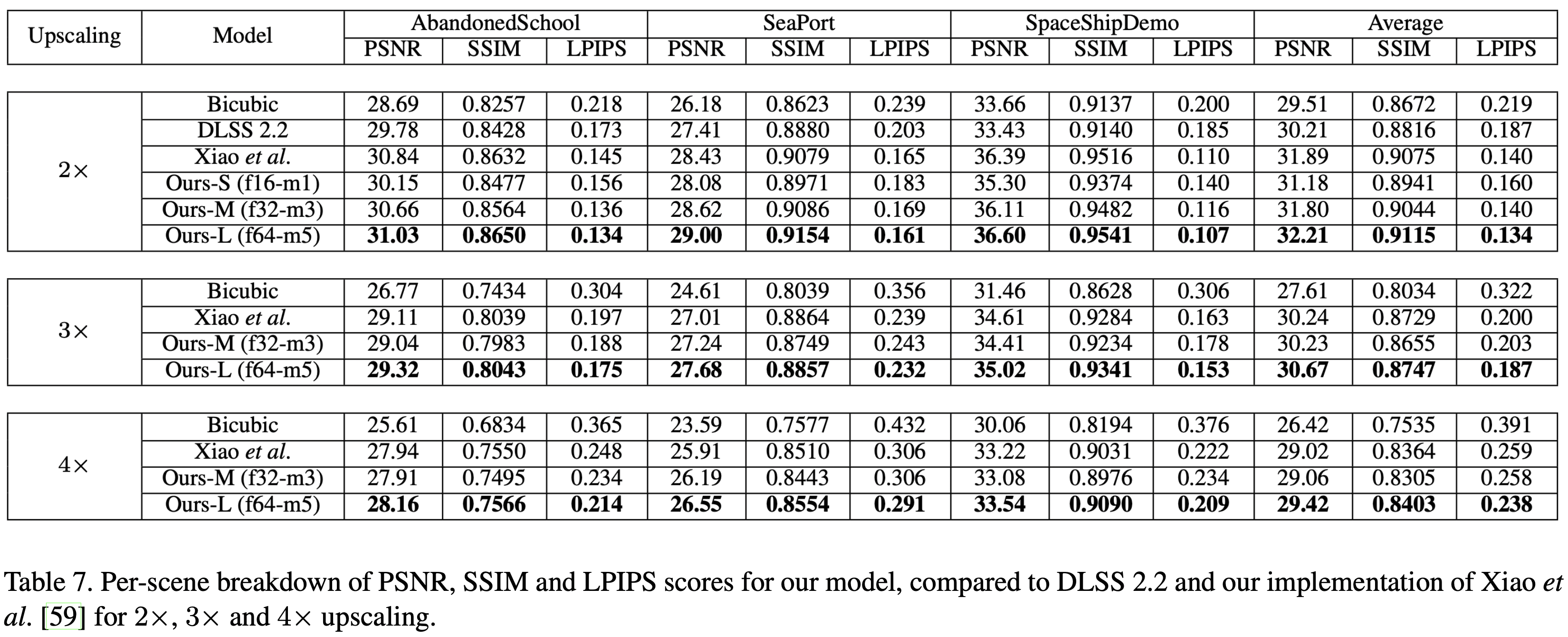}
\end{center}
\end{figure}

\end{document}